\documentclass{article}

\usepackage{microtype}
\usepackage{graphicx}
\usepackage{subcaption}
\usepackage{booktabs} % for professional tables
\usepackage{multirow}

\usepackage{hyperref}

\usepackage[preprint]{icml2026}

\usepackage{amsmath}
\usepackage{amssymb}
\usepackage{mathtools}
\usepackage{amsthm}

\usepackage[capitalize,noabbrev]{cleveref}

\theoremstyle{plain}

\theoremstyle{definition}

\theoremstyle{remark}

\usepackage[disable,textsize=tiny]{todonotes}

\icmltitlerunning{Visual-to-Symbolic Analytical Solution Inference from Field Visualizations}

\begin{document}

\twocolumn[
  \icmltitle{Hidden in Plain Sight: Visual-to-Symbolic\\
    Analytical Solution Inference from Field Visualizations}



  \icmlsetsymbol{equal}{*}

  \begin{icmlauthorlist}
    \icmlauthor{Pengze Li}{fudan,ailab}
    \icmlauthor{Jiaquan Zhang}{fudan}
    \icmlauthor{Yunbo Long}{cambridge}
    \icmlauthor{Xinping Liu}{ailab}\\
    \icmlauthor{Zhou wenjie}{nankai}
    \icmlauthor{Encheng Su}{ailab}
    \icmlauthor{Zihang Zeng}{fudan}
    \icmlauthor{Jiaqi Liu}{UNC}
    \icmlauthor{Jiyao Liu}{fudan}\\
    \icmlauthor{Junchi Yu}{oxford}
    \icmlauthor{Lihao Liu}{ailab}
    \icmlauthor{Philip Torr}{oxford}
    \icmlauthor{Shixiang Tang}{ailab}
    \icmlauthor{Aoran Wang}{ailab}
    \icmlauthor{Xi Chen}{fudan}
  \end{icmlauthorlist}

  \icmlaffiliation{fudan}{Fudan University, China}
  \icmlaffiliation{ailab}{Shanghai Artificial Intelligence Laboratory, China}
  \icmlaffiliation{cambridge}{University of Cambridge, UK}
  \icmlaffiliation{nankai}{Nankai University, China}
  \icmlaffiliation{UNC}{UNC-Chapel Hill, US}
  \icmlaffiliation{oxford}{University of Oxford, UK}

  \icmlcorrespondingauthor{Xi Chen}{x\_chen@fudan.edu.cn}
  \icmlcorrespondingauthor{Aoran Wang}{wangaoran@pjlab.org.cn}

  \icmlkeywords{Vision-Language Model, Symbolic Regression, Scientific Reasoning, Chain-of-Thought}

  \vskip 0.3in
]



\printAffiliationsAndNotice{}  % no special notice (required even if empty)

\begin{abstract}
Recovering analytical solutions of physical fields from visual observations is a fundamental yet underexplored capability for AI-assisted scientific reasoning.
We study \textbf{visual-to-symbolic analytical solution inference (ViSA)} for two-dimensional linear steady-state fields: given field visualizations (and first-order derivatives) plus minimal auxiliary metadata, the model must output a single executable \texttt{SymPy} expression with fully instantiated numeric constants.
We introduce \textbf{ViSA-R2} and align it with a self-verifying, solution-centric chain-of-thought pipeline that follows a physicist-like pathway: structural pattern recognition $\rightarrow$ solution-family (ansatz) hypothesis $\rightarrow$ parameter derivation $\rightarrow$ consistency verification.
We also release \textbf{ViSA-Bench}, a VLM-ready synthetic benchmark covering 30 linear steady-state scenarios with verifiable analytical/symbolic annotations, and evaluate predictions by numerical accuracy, expression-structure similarity, and character-level accuracy.
Using an 8B open-weight Qwen3-VL backbone, \textbf{ViSA-R2} outperforms strong open-source baselines and the evaluated closed-source frontier VLMs under a standardized protocol.
\end{abstract}

% Introduction

\section{Introduction}

\begin{figure*}[h]
    \centering
    \includegraphics[width=0.92\linewidth]{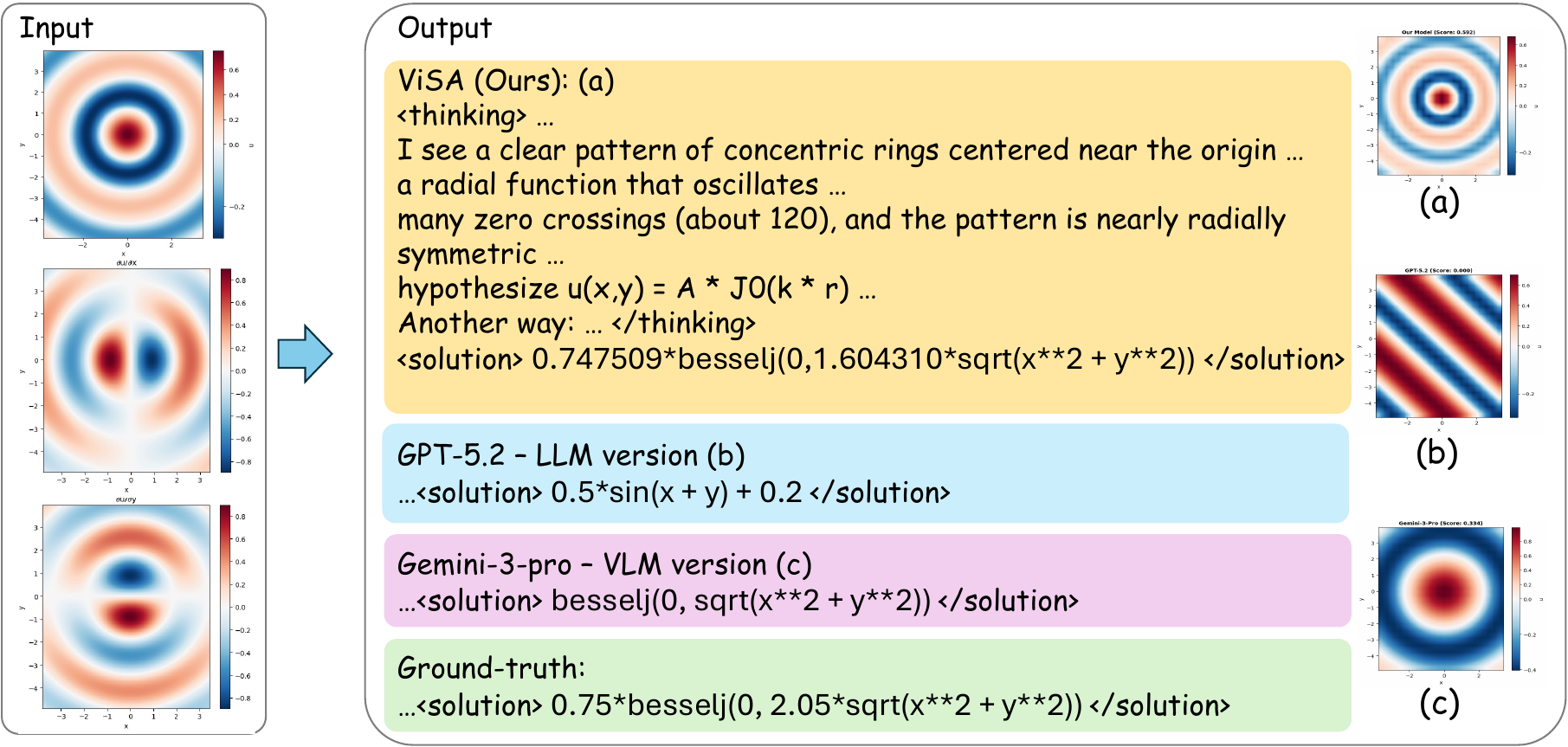}
    \caption{Visual-to-Symbolic Analytical Solution Inference}
    \label{fig:case}
\end{figure*}

Vision-language models (VLMs) have demonstrated strong capabilities in understanding the visual relationships encoded in various visual representations, including images, charts, and diagrams.
Such capability enables VLMs to identify trends, symmetries, and constraints in visual content, supporting a range of applications in diagram interpretation and data analysis~\cite{li2023blip2bootstrappinglanguageimagepretraining}.
However, understanding the visual relationship does not imply the ability to reason about the underlying rules that govern them.
For example, the VLM may correctly identify a curve in a diagram as sinusoidal in appearance, yet still struggle to identify the closed-form expression such as $A\sin(\omega x+\phi)+b$ with correct parameters.

This gap becomes significant in scientific workflows, where visual representations encode governing physical laws~\cite{lewkowycz2022solvingquantitativereasoningproblems,wang2024scibenchevaluatingcollegelevelscientific}.
Many physical quantities, such as temperature, electrostatic potential, concentration, or wave amplitude, are visualized as two-dimensional fields (e.g., heatmaps). 
Such visualizations expose visual signatures, such as symmetry, singularities, boundary-layer effects, decay and oscillation patterns, that strongly constrain the potential solution family. 
The key question we ask is: \emph{can a model use these visual signatures to infer an instance-specific closed-form solution, rather than producing a description or a purely numerical fit?}

In this work, we study \textbf{\underline{vi}sual-to-\underline{s}ymbolic \underline{a}nalytic solution inference (ViSA)} for two-dimensional linear steady-state fields in a controlled setting.
From one or more field visualizations (plus limited auxiliary numerical information), the model must produce a single executable \texttt{SymPy} expression whose numerical constants are fully instantiated, representing the solution for the given instance~\cite{gao2023palprogramaidedlanguagemodels,schick2023toolformerlanguagemodelsteach}.
We include text-only baselines that receive the same auxiliary numeric information but no images, allowing us to isolate the value of visual structure for symbolic inference. 
Our working hypothesis is that solution signatures like symmetry, singularities, and decay or oscillation patterns are hard to capture directly in token sequences, yet visually salient and consistent across instances.

To support rigorous evaluation, we construct a synthetic dataset (ViSA-Bench) covering 30 linear steady-state physics scenarios, with 500 parameterized instances per scenario.
The model's SymPy output is evaluated along three complementary axes: (i) numerical accuracy (ii) structure similarity (iii) character accuracy. Among these, structure similarity directly targets whether the model has inferred the correct solution form from visual patterns, distinguishing this task from pure numerical approximation.

To mitigate fitting without understanding, we synthesize high-quality chain-of-thought (CoT) trajectories for solution-centric training alignment. We extract visual cues from images, map them to ground-truth solution signatures, and derive key parameters with reproducible calculations. After fine-tuning, the model’s reasoning follows:
\emph{visual pattern recognition $\rightarrow$ solution-family (ansatz) hypothesis $\rightarrow$ parameter derivation $\rightarrow$ consistency verification}.
This setting differs from symbolic regression (SR) and LLM-only curve fitting, which mainly infer expressions from numeric samples and need not explicitly leverage visual structure.

Experiments show that CoT-aligned training yields consistent improvements across all three metrics and improves generalization across scenarios. Using an 8B open-weight Qwen3-VL backbone, \textbf{ViSA-R2} outperforms not only open-source baselines and the evaluated closed-source frontier VLMs under a standardized protocol. Together, these results support our central claim: leveraging visual field structures and explicit reasoning alignment provides an effective path toward end-to-end analytical solution inference.

We make three contributions:

\textbf{(1)}
We formulate \emph{visual-to-symbolic analytical solution inference} for 2D linear steady-state fields, and design evaluation settings for VLMs and LLMs to isolate the role of visual information.

\textbf{(2)}
We release a benchmark of 30 physically plausible scenarios with ground-truth analytical solutions, as well as  systematic evaluation of VLMs/LLMs on both symbolic recovery and numeric accuracy.

\textbf{(3)}
We propose a self-verifying pipeline to synthesize solution-centric CoT trajectories for training alignment. With CoT-aligned fine-tuning of an 8B Qwen3-VL backbone, our ViSA-R2 model achieves state-of-the-art performance and outperforms evaluated closed-source frontier VLMs under a standardized protocol.

We focus on linear steady-state fields as a controlled start; extending to nonlinear, time-dependent, and noisier observations is future work.

% Related Work

\begin{figure*}[t]
    \centering
    \includegraphics[width=0.9\linewidth]{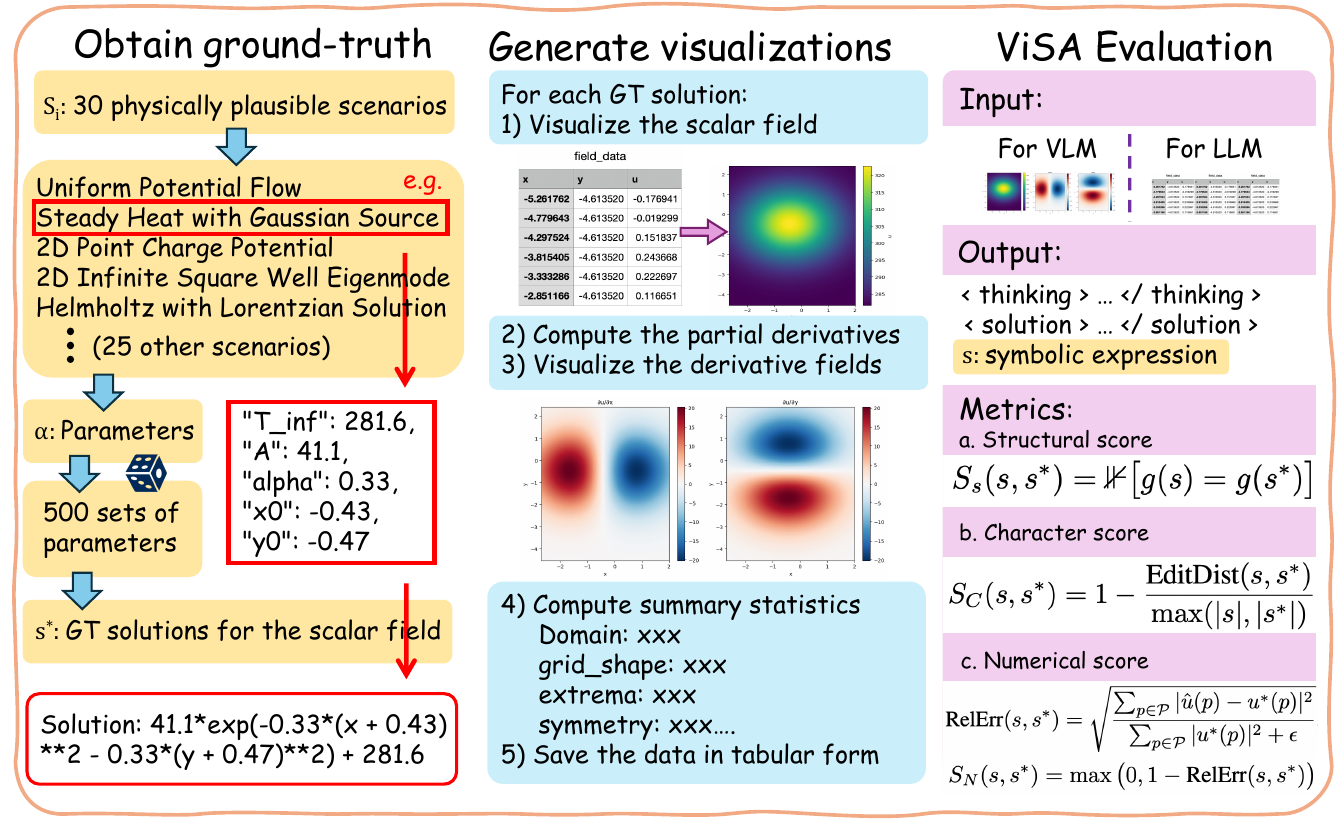}
    \caption{Overall pipeline for ViSA dataset construction and evaluation.
    }
    \label{fig:visa_pipeline}
\end{figure*}

\section{Related Work}
\subsection{Symbolic Regression and Equation Discovery}

Symbolic Regression (SR) seeks mathematical expressions from data, traditionally through genetic programming \cite{koza1992gp, schmidt2009science} and later via sparse regression (e.g., SINDy \cite{brunton2016sindy}) or physics-informed priors (e.g., AI Feynman \cite{udrescu2020aifeynman, udrescu2020aifeynman2}). These approaches focus on discovering governing equations from point sets or time series but are not designed to exploit the spatial-geometric patterns evident in field visualizations.
Recent LLM-assisted SR methods \cite{shojaee2025llmsr,zhang2025ragsr} treat equation discovery as program synthesis, using large language models to propose expression templates. However, these approaches still  require a large amount of sample points to approximate the field, which reduce the rich 2D field to tabular or sequential data—a inefficient transformation that discards the spatial structure that is immediately apparent in the image and essential for inferring the underlying solution form.
A few recent works have begun to bridge visual perception and symbolic reasoning for physics discovery. ViSymRe~\cite{li2024visymre} investigates vision-guided symbolic regression by coupling visual representations with a Transformer decoder to predict an equation skeleton. Notably, VIPER-R1~\cite{liu2025viper} combines VLM perception with symbolic modules to recover physical laws from motion trajectories. 
Our work introduces visual‑to‑symbolic solution inference and scales this challenge to 30 distinct linear steady‑state scenarios.

\subsection{Benchmarks for Multimodal Reasoning}

Existing multimodal evaluation suites such as MME~\cite{fu2023mme}, MMBench~\cite{liu2023mmbench}, and MM-Vet~\cite{yu2024mmvet} aim to assess broad VLM capabilities by disentangling perception from reasoning or grading open‑ended responses. Beyond such suites, PhysReason~\cite{zhang2025physreason},PhysUniBench~\cite{wang2025physunibench} and UGPhysics~\cite{xu2025ugphysics} provide different physics benchmark for diagram-grounded multimodal reasoning.  Complementary benchmarks like S‑Chain emphasize explanation fidelity for scientific imagery but remain confined to medical diagnosis scenarios~\cite{leduc2024schain}. These efforts, while valuable, largely address \emph{descriptive} or \emph{classificatory} tasks (e.g., question answering, captioning, or diagnosis) and rarely target the \emph{symbolic generative} challenge of inferring a compact mathematical expression from a visual observation.

Moreover, current multimodal scientific datasets are predominantly built around textual or tabular data, 1‑D curves, or simple diagrams—as seen in chart‑understanding benchmarks (DVQA, PlotQA, ChartQA),~\cite{kafle2018dvqa,methani2020plotqa,masry2022chartqa} diagram‑based QA (AI2D, TQA),~\cite{kembhavi2016ai2d,kembhavi2017tqa} and science QA with explanations~\cite{lu2022scienceqa,zhang2023mmcot, su2026sciif}. They lack the dense, spatially continuous representations characteristic of 2‑D steady‑state fields—the very setting where visual structure (symmetry, singularities, boundary layers) carries essential information about the underlying solution. More critically, most existing benchmarks provide only input‑output pairs without high‑quality, verifiable reasoning traces,~\cite{lu2024mathvista,yue2024mmmu} making it difficult to align models with the step‑wise, solution‑centric reasoning that human experts employ.
Our benchmark fills this gap with a VLM‑ready dataset of 30 linear steady‑state field scenarios, a three‑dimensional evaluation (numerical, structural, and character‑level), and a self‑verified pipeline for generating solution‑centric reasoning trajectories.

\subsection{VLMs for Mathematical and Scientific Reasoning}

While modern VLMs demonstrate impressive capabilities in visual parsing, captioning, and basic question answering, they struggle significantly when tasked with visual symbolic abstraction and formalization—the ability to distill continuous visual patterns into compact symbolic expressions.
Current VLMs excel at describing scientific visualizations: they can interpret charts, plots, and diagrams, extracting values and answering questions about their content~\cite{kafle2018dvqa,methani2020plotqa,masry2022chartqa,kembhavi2016ai2d,kembhavi2017tqa}. With chain-of-thought prompting, they can even generate textual rationales for science QA tasks~\cite{lu2022scienceqa,zhang2023mmcot}. However, these capabilities remain fundamentally descriptive—the model explains or answers questions about the data, but does not derive a formal mathematical representation from it.

This limitation becomes starkly evident in more demanding mathematical reasoning settings. Evaluations like MathVista and MMMU show that even state-of-the-art VLMs perform poorly on problems requiring symbolic abstraction and formalization~\cite{lu2024mathvista,yue2024mmmu}. The challenge is particularly acute for tasks that involve translating a continuous visual pattern (e.g., a heatmap's gradient, symmetry, or singularity structure) into a discrete, parameterized mathematical expression. While specialized models exist for mathematical notation recognition (e.g., image-to-LaTeX systems), these merely transcribe already-rendered equations rather than inferring new ones from observed phenomena~\cite{deng2016im2markup}.
Thus, there is a capability gap: VLMs lack the capacity to observe a physical field and produce a compact symbolic solution that encodes its structure.

\section{Methodology}
\label{sec:methodology}
\begin{figure*}[t]
    \centering
    \includegraphics[width=0.95\linewidth]{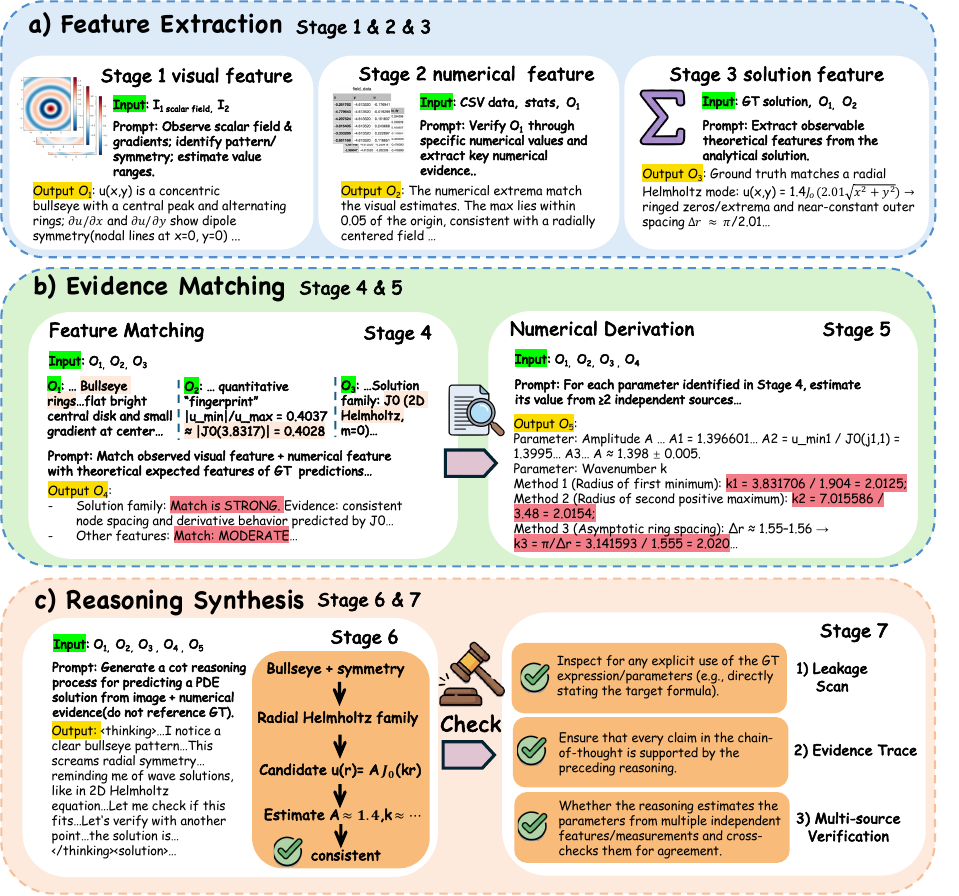}
    \caption{Overall pipeline for feature extraction, evidence matching, parameter inference, and synthesize high-quality reasoning chain.
    }
    \label{fig:cot_pipeline}
\end{figure*}

\subsection{Problem Setup}
\label{subsec:problem_setup}

We study \emph{visual-to-symbolic analytical solution inference} for 2D linear steady-state fields. Given an observation $\mathcal{X}$ that includes field visualizations (and optionally a small amount of auxiliary numerical information), the goal is to generate a SymPy expression $s$ that represents the underlying instance-specific solution $u(x,y)$.

\textbf{Input.}
We define $\mathcal{X}=(I_1,I_2,m)$. The visual input consists of two RGB heatmaps $I_1,I_2\in\mathbb{R}^{H\times W\times 3}$. Specifically, $I_1$ visualizes the field value $u(x,y)$ over a domain $\Omega\subset\mathbb{R}^2$. The second image $I_2$ encodes first-order derivatives: its left half visualizes $\partial u/\partial x$ and its right half visualizes $\partial u/\partial y$. In addition, $m$ denotes auxiliary numerical information such as domain bounds, normalization scales, and basic field statistics (e.g., mean and standard deviation). In our VLM setting, $m$ is intentionally minimal (typically fewer than 10 scalars). In the LLM-only baseline, images are removed and $m$ is expanded to include tabulated numeric samples.

\textbf{Output.}
The target output is a symbolic expression string $s\in\texttt{SymPyStrings}$ that conforms to SymPy syntax and contains explicit numeric parameters, representing the instance-specific solution. The expression may use standard arithmetic operations and common mathematical functions; in our synthetic data generation we draw expressions from a library $\mathcal{F}$ that includes elementary functions (e.g., \texttt{sin}, \texttt{cos}, \texttt{exp}, \texttt{log}) and special functions (e.g., \texttt{sinh}, \texttt{cosh}, \texttt{besselj}, \texttt{bessely}), while the model itself is not constrained by a hard decoding rule beyond producing valid SymPy.

\textbf{Task formulation.}
We model the mapping from observations to symbolic solutions as
\begin{equation}
\label{eq:task}
s = f_\theta(I_1, I_2, m), \qquad s \in \texttt{SymPyStrings},
\end{equation}
where $f_\theta$ is a vision--language model parameterized by $\theta$. When using chain-of-thought (CoT) alignment, the model additionally produces an intermediate rationale $r$ together with the final expression:
\begin{equation}
\label{eq:task_cot}
(r, s) = f_\theta(I_1, I_2, m).
\end{equation}
A training sample thus takes the form $(\mathcal{X}, r^*, s^*)$, where $s^*$ is the ground-truth SymPy expression and $r^*$ is a synthetically generated reasoning trace (see \S\ref{subsec:cot_synthesis}).

\subsection{Dataset Construction}
\label{subsec:dataset}

We construct a VLM-ready benchmark via scenario-based parametric generation. Each scenario defines a parametric family of closed-form solutions, allowing us to generate diverse field visual patterns while keeping ground-truth supervision symbolic and verifiable.

\textbf{Scenario definition.}
A scenario $\mathcal{S}_i$ is specified by (i) a solution template $f_i(x,y;\alpha)$ with parameters $\alpha\in\mathbb{R}^{d_i}$, and (ii) a parameter distribution $p_i(\alpha)$ that determines valid sampling ranges. For interpretability, some scenarios also provide a canonical linear steady-state operator form; when available, it enables an optional residual check, but the supervision signal throughout is the ground-truth solution expression.
To avoid numerical solver errors, we adopt a solution-first pipeline. For each instance, we sample $\alpha\sim p_i(\alpha)$ and instantiate the ground-truth field as $u^*(x,y)=f_i(x,y;\alpha)$ in SymPy. We then compute $\partial u^*/\partial x$ and $\partial u^*/\partial y$ by symbolic differentiation and evaluate them numerically on a grid. 

\textbf{Dataset coverage.}
We define 30 scenarios spanning visually distinguishable solution signatures, including logarithmic singularities (e.g., point-source potentials), smooth polynomial and exponential families, separable trigonometric--hyperbolic structures, oscillatory radial patterns (including Bessel-type modes), and selected special-function solutions (e.g., Airy-type). As a concrete example, a point-source potential scenario uses a logarithmic singularity template with parameters controlling source strength, scale, and singularity location.
For each of the 30 scenarios, we generate 500 parameterized instances, resulting in a pool of 15{,}000 samples. From this pool, we carefully curate a subset of 1{,}500 instances for gold CoT construction and CoT-aligned training, and we use an additional 150 instances for evaluation.

\subsection{Programmatic CoT Synthesis}
\label{subsec:cot_synthesis}

A key contribution is our \textit{programmatic CoT synthesis} framework for training-time reasoning alignment. The CoT is generated during dataset construction with access to the ground-truth expression.
For each instance, we are given observations $\mathcal{X}=(I_1,I_2,m)$ and the ground-truth SymPy expression $s^*$. We synthesize a rationale $r^*$ via a staged procedure. Crucially, Stages~1--2 operate only on $(I_1,I_2,m)$ to produce observation-based evidence, whereas later stages use $s^*$ to derive expected signatures and to construct a coherent derivation narrative, thereby reducing hallucinated rationales.

\textbf{Stages 1--3.}
Stage~1 extracts human-interpretable visual cues from $I_1$ and $I_2$, including approximate radial/axial symmetry, oscillatory versus monotone trends, sharp localized variations, and qualitative decay/growth patterns, without consulting $s^*$.
Stage~2 summarizes quantitative constraints from $m$ (e.g., domain bounds and basic statistics), also without consulting $s^*$.
Stage~3 then analyzes the ground-truth expression $s^*$ to derive the signatures it should induce in visualization and summary statistics (e.g., radial dependence via $\sqrt{x^2+y^2}$, periodicity from trigonometric terms, or logarithmic behavior near singularities).

\textbf{Stages 4--5.}
Stage~4 aligns the observation-based cues from Stages~1--2 with the expected signatures from Stage~3, and selects the subset of cues that most strongly supports the correct solution family.
Stage~5 constructs a step-by-step parameter-derivation narrative showing how the parameters in $s^*$ are consistent with the selected cues. The narrative references only observable quantities, but is generated with access to $s^*$ to ensure correctness.

\textbf{Stages 6--7.}
Stage~6 assembles the final rationale $r^*$ and outputs the final SymPy expression within a dedicated delimiter (e.g., \texttt{<solution>\dots</solution>}); in our gold-CoT set, the expression enclosed by \texttt{<solution>} is exactly $s^*$.
Stage~7 performs validation by (i) inspecting the rationale for any explicit use of the ground-truth expression or parameters (e.g., directly stating the target formula), (ii) ensuring that each claim in the chain-of-thought is supported by the preceding reasoning, and (iii) checking whether parameter estimates are derived from multiple independent features and cross-validated for agreement.

\begin{table*}[t]
\centering
\small
\setlength{\tabcolsep}{4pt}
\caption{\textbf{Overall comparison in VLM vs. LLM settings.}
We report the OverallScore and its three components for both VLM inputs and LLM-only inputs under the standardized protocol in \S\ref{subsec:inference}. We only include models that successfully return valid outputs for more than 90\% of the test instances.}

\label{tab:main_results}
\begin{tabular}{l|cccc|cccc}
\toprule
& \multicolumn{4}{c|}{\textbf{VLM}} & \multicolumn{4}{c}{\textbf{LLM-only}} \\
\cmidrule(lr){2-5}\cmidrule(lr){6-9}
\textbf{Model} & \textbf{Overall} & \textbf{Char} & \textbf{Struct} & \textbf{Num}
              & \textbf{Overall} & \textbf{Char} & \textbf{Struct} & \textbf{Num} \\
\midrule

GPT-5.2~\cite{openai_gpt52_2025} & 0.474 & 0.290 & 0.768 & 0.370 & 0.200 & 0.273 & 0.323 & 0.295 \\
Grok-4-1-Fast~\cite{xai_grok_4_1_2025} & 0.242 & 0.169 & 0.613 & 0.048 & 0.116 & 0.217 & 0.195 & 0.176 \\
Claude-Haiku-4-5\cite{anthropic_claude_haiku_4_5_2025} & 0.316 & 0.204 & 0.638 & 0.167 & 0.159 & 0.221 & 0.261 & 0.221 \\
\midrule
\textbf{ViSA-R2 (Ours)} & \textbf{0.512} & \textbf{0.482} & \textbf{0.860} & \textbf{0.385} & -- & -- & -- & -- \\

\bottomrule
\end{tabular}
\end{table*}

\section{Experiments}
\label{sec:experiments}

\subsection{Inference Protocol}
\label{subsec:inference}

At test time, ViSA-R2 generates a complete output sequence. We extract the predicted SymPy expression by parsing the content inside \texttt{<solution>...</solution>}. If the extracted string cannot be parsed by SymPy, we apply minimal normalization (e.g., whitespace cleanup). If parsing still fails, the prediction is marked invalid and assigned worst-case scores.

\textbf{Baseline comparisons.}
We compare against both VLM and LLM baselines under a standardized protocol. All models receive the same task instructions and the same required output format. VLM baselines receive $(I_1,I_2,m)$ as input. LLM-only baselines do not receive images; instead, they are provided with an expanded numeric table $m_{\text{expanded}}$ that contains 400 sampled tuples $(x,y,u,\partial u/\partial x,\partial u/\partial y)$. We evaluate baselines with and without test-time CoT prompting, to separate the effect of prompting from training-time CoT alignment.

\label{subsec:metrics}

We evaluate predicted expressions with three complementary metrics that capture syntactic fidelity, structural correctness, and numeric accuracy.

\textbf{Character-level accuracy.}
We measure string-level similarity using a normalized edit-distance score,
\begin{equation}
\label{eq:char_score}
\text{$S_C$}(s, s^*) = 1 - \frac{\text{EditDist}(s, s^*)}{\max(|s|, |s^*|)},
\end{equation}
where $\text{EditDist}(\cdot,\cdot)$ is the Levenshtein distance and $|\cdot|$ denotes string length.

\textbf{Expression-structure similarity.}
To assess whether the model recovers the correct functional form independent of numeric precision, we define a structure-normalization operator $g(\cdot)$ that (i) parses the expression into a SymPy AST when possible, (ii) simplifies it, and (iii) replaces all numeric constants with a shared placeholder symbol $\texttt{C}$. The structural score is then a binary match:
\begin{equation}
\label{eq:struct_score}
\text{$S_s$}(s, s^*) = \mathbb{1}\big[g(s)=g(s^*)\big].
\end{equation}

\textbf{Numerical accuracy.}
We evaluate the predicted expression by executing it to obtain $\hat{u}(\cdot)$ and computing a relative $L^2$ error on an evaluation set of points $\mathcal{P}\subset\Omega$,
\begin{equation}
\label{eq:rel_err}
\text{RelErr}(s, s^*) =
\sqrt{\frac{\sum_{p \in \mathcal{P}} |\hat{u}(p) - u^*(p)|^2}
{\sum_{p \in \mathcal{P}} |u^*(p)|^2 + \epsilon}},
\end{equation}
with $\epsilon=10^{-8}$. We convert this to a bounded score by
\begin{equation}
\label{eq:num_score}
\text{$S_N$}(s, s^*) = \max\big(0, 1 - \text{RelErr}(s, s^*)\big).
\end{equation}
If $s$ fails to parse or raises runtime errors during evaluation, we set $\text{NumScore}=0$.

\textbf{Overall score.}
For a single summary statistic, we report a weighted geometric mean:
\begin{equation}
\label{eq:overall_score}
\text{OverallScore}(s,s^*) =
\text{$S_C$}^{0.2}\cdot \text{$S_S$}^{0.3}\cdot \text{$S_N$}^{0.5}.
\end{equation}

\subsection{Model and Training}
\label{subsec:training}

We develop ViSA-R2 by fine-tuning a pretrained vision--language model with native multi-image conditioning.
The model generates a single text sequence that contains an intermediate rationale (CoT) and a final SymPy expression enclosed in \texttt{<solution>...</solution>} delimiters. We fine-tune a Qwen3-VL model (8B) \cite{qwen3} using bfloat16 precision with a learning rate of $2\times10^{-5}$ for 4 epochs. We use a maximum sequence length of 8192 tokens, per-device batch size of 1, and gradient accumulation over 4 steps. 

\textbf{Supervised fine-tuning objective.}
We fine-tune on triplets $(\mathcal{X}, r^*, s^*) \sim \mathcal{D}$ using standard next-token prediction, where $\mathcal{X}$ denotes the input context (e.g., the prompt and any observations). Let $y^* = [r^*; s^*]$ be the concatenated target sequence with length $T = |y^*|$. The training loss is
\begin{equation}
\label{eq:sft_loss}
\mathcal{L}(\theta) = -\mathbb{E}_{(\mathcal{X}, r^*, s^*) \sim \mathcal{D}}
\left[\sum_{t=1}^{T} \log p_\theta\!\left(y_t^* \mid y_{<t}^*, \mathcal{X}\right) \right],
\end{equation}
where $p_\theta$ is the model's conditional token distribution.
We supervise both the synthesized rationale $r^*$ and the final expression $s^*$ by training on the concatenated sequence $[r^*; s^*]$, thereby enabling the model to learn to generate $r^*$ followed by $s^*$ under the same token-level negative log-likelihood.

\subsection{Main Results}
\label{subsec:main_results}

Table~\ref{tab:main_results} summarizes the main comparison. Across LLM baselines, the dominant failure mode is incorrect structure prediction. ViSA-R2 achieves the best overall performance on our benchmark, with particularly strong structure recovery, indicating that the model reliably identifies the visually implied functional form. This supports our central claim that leveraging field and gradient visualizations with training-time CoT alignment is effective for visual-to-symbolic analytical solution inference.

We evaluated additional models beyond Table~\ref{tab:main_results}, including Gemini-3.0-Pro~\cite{google_gemini_3_pro_2025}, Grok-4~\cite{xai_grok_4_2025}, Claude-Sonnet-4~\cite{anthropic_claude_sonnet_4_2025}, and DeepSeek-R1~\cite{deepseek_r1_2025}. Under our batched evaluation pipeline, these models frequently yielded empty, truncated, or non-parsable outputs (i.e., failing to adhere to the required \texttt{<solution>} format). This instability persisted even with a larger context budget (up to 40k tokens) and prompt variants (e.g., explicitly constraining the budget for intermediate reasoning). Consequently, we only obtained results on a subset of instances. Since the evaluated subsets are not identical, including these numbers in the main table would preclude a fair comparison; we therefore report them seperately in~\ref{tab:vlm_llm_benchmark} as coarse, non-comparable trends. More broadly, this highlights that stable, parsable generation is a practical prerequisite for reliable batched evaluation.

\begin{table}[t]
  \centering
  \small
  \setlength{\tabcolsep}{0pt}
    \caption{Comparison of representative models by overall score and success rate.}
  \label{tab:vlm_llm_benchmark}
  \begin{tabular*}{\linewidth}{@{\extracolsep{\fill}}lcccc}
  \toprule
  \textbf{Model} & \textbf{Input} & \textbf{Overall} & \textbf{Success}\\
  \midrule
  ViSA-R2 framework & Vision & 0.512 & 100.0\% \\
  Claude-Haiku-4-5~\cite{anthropic_claude_haiku_4_5_2025} & Vision  & 0.316 & 93.3\% \\
  Claude Sonnet 4.5~\cite{anthropic_claude_sonnet_4_2025} & Text  & 0.352 & 63.3\% \\
  Gemini-3-Pro~\cite{google_gemini_3_pro_2025} & Vision & 0.572 & 60.0\% \\
  Claude Opus 4 ~\cite{anthropic_claude_opus_4_2025}& Text & 0.135 & 26.7\% \\
  DeepSeek-R1~\cite{deepseek_r1_2025} & Text & 0.915 & 10.0\% \\
  Grok-4~\cite{xai_grok_4_2025} & Text & 0.828 & 3.3\% \\

  \bottomrule
  \end{tabular*}
\end{table}

\textbf{Ablation A: VLM vs.\ LLM-only inputs.}
This ablation tests whether visual structure provides signal beyond tabulated numeric samples. The VLM setting uses $(I_1, I_2, m)$, while the LLM-only setting removes images and instead provides an expanded numeric table $m_{\text{expanded}}$ consisting of sampled tuples $(x, y, u, \partial u/\partial x, \partial u/\partial y)$. Table~\ref{tab:main_results} reports the overall score and its three components for both settings.

Across all evaluated models, performance is consistently higher in the VLM setting than in the LLM-only setting. The largest gain is observed on structure for GPT-5.2~\cite{openai_gpt52_2025}, which increases from 0.323 to 0.768. This suggests that, in symbolic-regression-style workflows, VLMs exhibit a strong visual inductive bias: they can recover substantially more structural information from visual cues than from text-only numeric tables.

\textbf{Ablation B: training-time CoT alignment vs.\ parameter optimization.}
This ablation tests whether the observed gains primarily come from training-time reasoning alignment, rather than test-time prompting. We compare ViSA-R2 fine-tuned with gold chain-of-thought (CoT) supervision to an otherwise identical model trained to output only the final solution expression (i.e., no intermediate reasoning). At inference time, both variants are evaluated under the same standardized protocol and the same output format. Without CoT alignment during training, ViSA-R2 frequently exhibits repetitive, looping reasoning and often fails to produce a valid solution output. 
Invalid or missing outputs are assigned a score of 0, which substantially lowers the scores of the w/o CoT alignment variant. As an additional control, we also include a test-time CoT prompting condition, where the model is prompted to generate CoT at inference without any CoT supervision during training. This setting can be viewed as a no-training baseline for reasoning elicitation.

Additionally, since accurately regressing continuous coefficients is not the primary strength of VLM-based symbolic generation, we denote ViSA-R2 framework as ViSA-R2 augmented with coefficient refinement. Concretely, we augment ViSA-R2 with a lightweight coefficient-refinement stage (L-BFGS-B), minimizing the mean squared error to target field values.
This hybrid variant isolates coefficient fitting from structure discovery and substantially improves \text{$S_N$} with minimal overhead.

\begin{table}[t]
\centering
\small
\setlength{\tabcolsep}{4pt}
\caption{\textbf{Ablation B.} CoT alignment and coefficient refinement.}
\label{tab:abl_cot}
\begin{tabular}{lcccc}
\toprule
\textbf{Variant}  & \textbf{Char} & \textbf{Struct} & \textbf{Num} & \textbf{Overall}\\
\midrule
Qwen-3-8B-baseline & 0.191 & 0.616 & 0.210 & 0.328 \\
w/o CoT alignment & 0.016 & 0.053 & 0.017 & 0.024 \\
w/ CoT alignment (ViSA-R2) & 0.482 & 0.860 & 0.385 & 0.512 \\
ViSA-R2 framework & 0.480 & 0.842 & 0.780 & 0.669 \\
\bottomrule
\end{tabular}
\end{table}
\label{subsec:ablation}

\section{Discussion}
\label{sec:discussion}

\subsection{Revisiting the Core Question}
\label{subsec:disc_core}

This paper investigated whether a VLM can translate field visual observations into an instance-specific symbolic analytical solution, rather than producing descriptive statements or purely numerical fits. Under our benchmark and standardized protocol, the results suggest an affirmative answer. ViSA-R2 achieves the best overall performance and substantially improves expression-structure similarity, indicating reliable recovery of the correct functional form from visual cues.

The post-hoc parameter refinement study further disentangles structure from coefficients: once the predicted structure is correct, fitting coefficients on the available sampled observations can often reduce numerical error substantially. This supports a practical workflow in which the VLM proposes candidate symbolic forms, and lightweight numerical routines calibrate constants with minimal additional overhead.

Our ablations reinforce this interpretation. VLM inputs outperform baselines using LLM-only numeric tables under the same protocol, consistent with the hypothesis that key solution signatures are visually salient and informative for navigating the solution space. Moreover, training-time CoT alignment improves structural correctness, suggesting that explicit reasoning supervision helps the model internalize the intermediate steps needed for reliable structure recovery. Collectively, these findings highlight visual structure as a compact and effective inductive signal for symbolic recovery.

\subsection{Limitations and What Did Not Work}
\label{subsec:disc_limit}

\textbf{Why not annotate gold-CoT for all 15k instances?}

High-quality reasoning supervision is resource-intensive. We curated 1{,}500 gold-CoT instances because each rationale requires structured cue extraction, verification against ground-truth signatures, and a coherent parameter-derivation narrative. We attempted to scale supervision by prompting models to imitate or rewrite gold rationales for additional instances from similar scenarios. In practice, these synthetic rationales were not sufficiently reliable at scale, exhibiting hallucinated cues, inconsistent logic, and reasoning gaps that were difficult to filter. This finding suggests that solution-centric reasoning traces are not a trivial byproduct of prompting, and that structured synthesis remains indispensable for stable supervision.

\textbf{What constrained our API model choice for CoT construction?}

In preliminary trials, strong closed-source models (especially GPT-5 and Gemini-3-pro) often returned abbreviated explanations instead of full step-by-step rationales (e.g., omitting detailed chain-of-thought and offering only a brief summary). In our CoT workflow, we therefore used GPT-5 for single-step observation and cue articulation, and Grok-4 for compressing long traces into a consistent final rationale format, thereby improving stability and reducing formatting failures.

\subsection{Future Directions}
\label{subsec:disc_future}

Our goal is not to cover all scientific fields, but to establish a new task and a reliable evaluation protocol for visual-to-symbolic solution inference. Several immediate directions for future research emerge. First, we plan to extend beyond the current steady-state setting to time-dependent fields, where models must reason over sequences of observations. Second, we will expand operator families and regimes, including broader nonlinear systems and more complex boundary and source configurations. Third, we will study more realistic observation conditions, such as measurement noise, partial observability, and rendering artifacts, to better approximate practical scientific workflows. Finally, reducing the cost of reliable reasoning supervision remains central; we will explore stronger automated cue extractors and more effective filtering criteria to scale gold-CoT construction without sacrificing quality.

\section{Conclusion}
\label{sec:conclusion}

We have introduced the visual-to-symbolic analytical solution inference as a new VLM-centric capability: recovering an instance-specific SymPy expression (structure and numeric parameters) directly from visualizations of physical fields and their derivatives. To make this setting measurable and reproducible, we constructed a VLM-ready benchmark spanning 30 parametric steady-state scenarios with verifiable ground truth, and evaluated predictions using complementary structural, numerical, and character-level metrics. We further proposed a programmatic framework for synthesizing solution-centric CoT traces and applied CoT alignment during fine-tuning, enabling a single model, \textbf{ViSA-R2}, to reliably infer solution families and improve parameter estimation. Under a standardized protocol, ViSA-R2 outperforms strong open-source baselines and representative closed-source frontier VLMs, providing evidence that visualizations of fields and their derivatives, together with training-time reasoning alignment, form an effective pathway for end-to-end analytical solution inference. We view this work as providing a problem formulation and a practical paradigm, and we hope it motivates future research that extends visual-to-symbolic reasoning to broader operators, dynamics, and more realistic observation conditions.

\section*{Impact Statements}
This paper presents work whose goal is to advance the field of machine learning. There are many potential societal consequences of our work, none of which we feel must be specifically highlighted here.

\bibliography{example_paper}

@book{koza1992gp,
  title={Genetic Programming: On the Programming of Computers by Means of Natural Selection},
  author={Koza, John R.},
  publisher={MIT Press},
  year={1992},
  isbn={9780262527910},
  url={https://mitpress.mit.edu/9780262527910/genetic-programming/}
}

@article{schmidt2009science,
  title={Distilling Free-Form Natural Laws from Experimental Data},
  author={Schmidt, Michael and Lipson, Hod},
  journal={Science},
  year={2009},
  volume={324},
  number={5923},
  pages={81--85},
  doi={10.1126/science.1165893},
  url={https://pubmed.ncbi.nlm.nih.gov/19342586/}
}

@article{brunton2016sindy,
  title={Discovering governing equations from data by sparse identification of nonlinear dynamical systems},
  author={Brunton, Steven L. and Proctor, Joshua L. and Kutz, J. Nathan},
  journal={Proceedings of the National Academy of Sciences},
  year={2016},
  volume={113},
  number={15},
  pages={3932--3937},
  doi={10.1073/pnas.1517384113},
  url={https://pmc.ncbi.nlm.nih.gov/articles/PMC4839439/}
}

@article{udrescu2020aifeynman,
  title={{AI Feynman}: A physics-inspired method for symbolic regression},
  author={Udrescu, Silviu-Marian and Tegmark, Max},
  journal={Science Advances},
  year={2020},
  volume={6},
  number={16},
  pages={eaay2631},
  doi={10.1126/sciadv.aay2631},
  url={https://pubmed.ncbi.nlm.nih.gov/32426452/}
}

@misc{udrescu2020aifeynman2,
  title={{AI Feynman} 2.0: Pareto-optimal symbolic regression exploiting graph modularity},
  author={Udrescu, Silviu-Marian and Tan, Andrew and Feng, Jiahai and Neto, Orisvaldo and Wu, Tailin and Tegmark, Max},
  year={2020},
  howpublished={arXiv:2006.10782},
  url={https://arxiv.org/abs/2006.10782}
}

@inproceedings{shojaee2025llmsr,
  title={{LLM-SR}: Scientific Equation Discovery via Programming with Large Language Models},
  author={Shojaee, Parshin and Meidani, Kazem and Gupta, Shashank and Barati Farimani, Amir and Reddy, Chandan K.},
  booktitle={International Conference on Learning Representations (ICLR)},
  year={2025},
}

@inproceedings{zhang2025ragsr,
  title={{RAG-SR}: Retrieval-Augmented Generation for Neural Symbolic Regression},
  author={Zhang, Hengzhe and Chen, Qi and Xue, Bing and Banzhaf, Wolfgang and Zhang, Mengjie},
  booktitle={International Conference on Learning Representations (ICLR)},
  year={2025},
}

@misc{liu2025viper,
  title={Mimicking the Physicist's Eye: A VLM-centric Approach for Physics Formula Discovery},
  author={Liu, Jiaqi and Lai, Songning and Li, Pengze and Yu, Di and Zhou, Wenjie and Zhou, Yiyang and Xia, Peng and Wang, Zijun and Chen, Xi and Tang, Shixiang and Bai, Lei and Ouyang, Wanli and Ding, Mingyu and Yao, Huaxiu and Wang, Aoran},
  year={2025},
  howpublished={arXiv:2508.17380},
  url={https://arxiv.org/abs/2508.17380},
  note={Project page: https://jiaaqiliu.github.io/VIPER-R1/}
}

@inproceedings{kafle2018dvqa,
  title={{DVQA}: Understanding Data Visualizations via Question Answering},
  author={Kafle, Kushal and Price, Brian and Cohen, Scott and Kanan, Christopher},
  booktitle={Proceedings of the IEEE Conference on Computer Vision and Pattern Recognition (CVPR)},
  year={2018},
  pages={5648--5656},
  url={https://openaccess.thecvf.com/content_cvpr_2018/html/Kafle_DVQA_Understanding_Data_CVPR_2018_paper.html}
}

@inproceedings{methani2020plotqa,
  title={{PlotQA}: Reasoning over Scientific Plots},
  author={Methani, Nitesh and Ganguly, Pritha and Khapra, Mitesh M. and Kumar, Pratyush},
  booktitle={Proceedings of the IEEE/CVF Winter Conference on Applications of Computer Vision (WACV)},
  year={2020},
  pages={1527--1536},
  url={https://openaccess.thecvf.com/content_WACV_2020/html/Methani_PlotQA_Reasoning_over_Scientific_Plots_WACV_2020_paper.html}
}

@inproceedings{masry2022chartqa,
  title={{ChartQA}: A Benchmark for Question Answering about Charts with Visual and Logical Reasoning},
  author={Masry, Ahmed and Long, Do Xuan and Tan, Jia Qing and Joty, Shafiq and Hoque, Enamul},
  booktitle={Findings of the Association for Computational Linguistics: ACL 2022},
  year={2022},
  pages={2263--2279},
  doi={10.18653/v1/2022.findings-acl.177},
  url={https://aclanthology.org/2022.findings-acl.177/}
}

@inproceedings{kembhavi2016ai2d,
  title={A Diagram Is Worth A Dozen Images},
  author={Kembhavi, Aniruddha and Salvato, Mike and Kolve, Eric and Seo, Minjoon and Hajishirzi, Hannaneh and Farhadi, Ali},
  booktitle={Computer Vision -- ECCV 2016},
  year={2016},
  pages={235--251},
  series={Lecture Notes in Computer Science},
  volume={9908},
  publisher={Springer International Publishing},
  doi={10.1007/978-3-319-46493-0_15},
  url={https://doi.org/10.1007/978-3-319-46493-0_15}
}

@inproceedings{kembhavi2017tqa,
  title={Are You Smarter Than a Sixth Grader? Textbook Question Answering for Multimodal Machine Comprehension},
  author={Kembhavi, Aniruddha and Seo, Minjoon and Schwenk, Dustin and Choi, Jonghyun and Farhadi, Ali and Hajishirzi, Hannaneh},
  booktitle={Proceedings of the IEEE Conference on Computer Vision and Pattern Recognition (CVPR)},
  year={2017},
  pages={4999--5007},
  url={https://openaccess.thecvf.com/content_cvpr_2017/html/Kembhavi_Are_You_Smarter_CVPR_2017_paper.html}
}

@inproceedings{lu2022scienceqa,
  title={Learn to Explain: Multimodal Reasoning via Thought Chains for Science Question Answering},
  author={Lu, Pan and Mishra, Swaroop and Xia, Tony and Qiu, Liang and Chang, Kai-Wei and Zhu, Song-Chun and Tafjord, Oyvind and Clark, Peter and Kalyan, Ashwin},
  booktitle={Advances in Neural Information Processing Systems (NeurIPS)},
  year={2022},
}

@article{zhang2023mmcot,
  title={Multimodal Chain-of-Thought Reasoning in Language Models},
  author={Zhang, Zhuosheng and Zhang, Aston and Li, Mu and Zhao, Hai and Karypis, George and Smola, Alex},
  journal={Transactions on Machine Learning Research},
  year={2024},
  url={https://openreview.net/forum?id=y1pPWFVfvR}
}

@inproceedings{lu2024mathvista,
  title={MathVista: Evaluating Mathematical Reasoning of Foundation Models in Visual Contexts},
  author={Lu, Pan and Bansal, Hritik and Xia, Tony and Liu, Jiacheng and Li, Chunyuan and Hajishirzi, Hannaneh and Cheng, Hao and Chang, Kai-Wei and Galley, Michel and Gao, Jianfeng},
  booktitle={International Conference on Learning Representations (ICLR)},
  year={2024},
  url={https://iclr.cc/virtual/2024/oral/19768}
}

@inproceedings{yue2024mmmu,
  title={{MMMU}: A Massive Multi-discipline Multimodal Understanding and Reasoning Benchmark for Expert AGI},
  author={Yue, Xiang and Ni, Yuansheng and Zhang, Kai and Zheng, Tianyu and Liu, Ruoqi and Zhang, Ge and Stevens, Samuel and Jiang, Dongfu and Ren, Weiming and Sun, Yuxuan and Wei, Cong and Yu, Botao and Yuan, Ruibin and Sun, Renliang and Yin, Ming and Zheng, Boyuan and Yang, Zhenzhu and Liu, Yibo and Huang, Wenhao and Sun, Huan and Su, Yu and Chen, Wenhu},
  booktitle={Proceedings of the IEEE/CVF Conference on Computer Vision and Pattern Recognition (CVPR)},
  year={2024},
  pages={9556--9567},
  url={https://openaccess.thecvf.com/content/CVPR2024/html/Yue_MMMU_A_Massive_Multi-discipline_Multimodal_Understanding_and_Reasoning_Benchmark_for_CVPR_2024_paper.html}
}

@inproceedings{deng2016im2markup,
  title={Image-to-Markup Generation with Coarse-to-Fine Attention},
  author={Deng, Yuntian and Kanervisto, Anssi and Ling, Jeffrey and Rush, Alexander M.},
  booktitle={Proceedings of the 34th International Conference on Machine Learning},
  series={Proceedings of Machine Learning Research},
  volume={70},
  pages={980--989},
  year={2017},
  publisher={PMLR},
  url={https://proceedings.mlr.press/v70/deng17a.html}
}

@inproceedings{fu2023mme,
  title={{MME}: A Comprehensive Evaluation Benchmark for Multimodal Large Language Models},
  author={Fu, Chaoyou and Chen, Peixian and Shen, Yunhang and Qin, Yulei and Zhang, Mengdan and Lin, Xu and Yang, Jinrui and Zheng, Xiawu and Li, Ke and Sun, Xing and Wu, Yunsheng and Ji, Rongrong and Shan, Caifeng and He, Ran},
  booktitle={NeurIPS 2025 Datasets and Benchmarks Track},
  year={2025},
  url={https://openreview.net/forum?id=DgH9YCsqWm},
  note={Spotlight. arXiv:2306.13394}
}

@inproceedings{liu2023mmbench,
  title={{MMBench}: Is Your Multi-modal Model an All-Around Player?},
  author={Liu, Yuan and Duan, Haodong and Zhang, Yuanhan and Li, Bo and Zhang, Songyang and Zhao, Wangbo and Yuan, Yike and Wang, Jiaqi and He, Conghui and Liu, Ziwei and Chen, Kai and Lin, Dahua},
  booktitle={Computer Vision -- ECCV 2024},
  year={2024},
  pages={216--233},
  series={Lecture Notes in Computer Science},
  publisher={Springer Nature Switzerland},
  doi={10.1007/978-3-031-72658-3_13},
  url={https://doi.org/10.1007/978-3-031-72658-3_13}
}

@inproceedings{yu2024mmvet,
  title={{MM-Vet}: Evaluating Large Multimodal Models for Integrated Capabilities},
  author={Yu, Weihao and Yang, Zhengyuan and Li, Linjie and Wang, Jianfeng and Lin, Kevin and Liu, Zicheng and Wang, Xinchao and Wang, Lijuan},
  booktitle={International Conference on Machine Learning (ICML)},
  year={2024},
  url={https://icml.cc/virtual/2024/poster/34344}
}

@article{li2024visymre,
  title={Visymre: Vision-guided multimodal symbolic regression},
  author={Li, Da and Yin, Junping and Xu, Jin and Li, Xinxin and Zhang, Juan},
  journal={arXiv preprint arXiv:2412.11139},
  year={2024},
  url={https://arxiv.org/abs/2412.11139}
}

@misc{leduc2024schain,
       title={S-Chain: Structured Visual Chain-of-Thought for Medicine}, 
       author={Khai Le-Duc and Phuong T. H. Trinh and Duy M. H. Nguyen and Tien-Phat Nguyen and Nghiem T. Diep and An Ngo and Tung Vu and Trinh Vuong and Anh-Tien Nguyen and Mau Nguyen and Van Trung Hoang and Khai-Nguyen Nguyen and Hy Nguyen and Chris Ngo and Anji Liu and Nhat Ho and Anne-Christin Hauschild and Khanh Xuan Nguyen and Thanh Nguyen-Tang and Pengtao Xie and Daniel Sonntag and James Zou and Mathias Niepert and Anh Totti Nguyen},
       year={2025},
       howpublished={arXiv:2510.22728},
       url={https://arxiv.org/abs/2510.22728},
       note={Project page: https://s-chain.github.io/}
}

@misc{su2026sciif,
       title={SciIF: Benchmarking Scientific Instruction Following Towards Rigorous Scientific Intelligence}, 
       author={Encheng Su and Jianyu Wu and Chen Tang and Lintao Wang and Pengze Li and Aoran Wang and Jinouwen Zhang and Yizhou Wang and Yuan Meng and Xinzhu Ma and Shixiang Tang and Houqiang Li},
       year={2026},
       howpublished={arXiv:2601.04770},
       url={https://arxiv.org/abs/2601.04770}, 
}

@misc{wang2025physunibench,
       title={PhysUniBench: An Undergraduate-Level Physics Reasoning Benchmark for Multimodal Models}, 
       author={Lintao Wang and Encheng Su and Jiaqi Liu and Pengze Li and Peng Xia and Jiabei Xiao and Wenlong Zhang and Xinnan Dai and Xi Chen and Yuan Meng and Mingyu Ding and Lei Bai and Wanli Ouyang and Shixiang Tang and Aoran Wang and Xinzhu Ma},
       year={2025},
       howpublished={arXiv:2506.17667},
       url={https://arxiv.org/abs/2506.17667}, 
}

@inproceedings{gao2023palprogramaidedlanguagemodels,
  title={{PAL}: Program-aided Language Models},
  author={Gao, Luyu and Madaan, Aman and Zhou, Shuyan and Alon, Uri and Liu, Pengfei and Yang, Yiming and Callan, Jamie and Neubig, Graham},
  booktitle={Proceedings of the 40th International Conference on Machine Learning},
  series={Proceedings of Machine Learning Research},
  volume={202},
  pages={10764--10799},
  year={2023},
  publisher={PMLR},
  url={https://proceedings.mlr.press/v202/gao23f.html}
}

@inproceedings{schick2023toolformerlanguagemodelsteach,
  title={Toolformer: Language Models Can Teach Themselves to Use Tools},
  author={Schick, Timo and Dwivedi-Yu, Jane and Dessi, Roberto and Raileanu, Roberta and Lomeli, Maria and Hambro, Eric and Zettlemoyer, Luke and Cancedda, Nicola and Scialom, Thomas},
  booktitle={Advances in Neural Information Processing Systems},
  volume={36},
  pages={68539--68551},
  year={2023},
  publisher={Curran Associates, Inc.},
}

@inproceedings{li2023blip2bootstrappinglanguageimagepretraining,
  title={{BLIP}-2: Bootstrapping Language-Image Pre-training with Frozen Image Encoders and Large Language Models},
  author={Li, Junnan and Li, Dongxu and Savarese, Silvio and Hoi, Steven},
  booktitle={Proceedings of the 40th International Conference on Machine Learning},
  series={Proceedings of Machine Learning Research},
  volume={202},
  pages={19730--19742},
  year={2023},
  publisher={PMLR},
  url={https://proceedings.mlr.press/v202/li23q.html}
}

@inproceedings{lewkowycz2022solvingquantitativereasoningproblems,
  title={Solving Quantitative Reasoning Problems with Language Models},
  author={Lewkowycz, Aitor and Andreassen, Anders and Dohan, David and Dyer, Ethan and Michalewski, Henryk and Ramasesh, Vinay and Slone, Ambrose and Anil, Cem and Schlag, Imanol and Gutman-Solo, Theo and Wu, Yuhuai and Neyshabur, Behnam and Gur-Ari, Guy and Misra, Vedant},
  booktitle={Advances in Neural Information Processing Systems},
  volume={35},
  pages={3843--3857},
  year={2022},
  publisher={Curran Associates, Inc.},
  
}

@inproceedings{wang2024scibenchevaluatingcollegelevelscientific,
  title={{SciBench}: Evaluating College-Level Scientific Problem-Solving Abilities of Large Language Models},
  author={Wang, Xiaoxuan and Hu, Ziniu and Lu, Pan and Zhu, Yanqiao and Zhang, Jieyu and Subramaniam, Satyen and Loomba, Arjun R and Zhang, Shichang and Sun, Yizhou and Wang, Wei},
  booktitle={Proceedings of the 41st International Conference on Machine Learning},
  series={Proceedings of Machine Learning Research},
  volume={235},
  pages={50622--50649},
  year={2024},
  publisher={PMLR},
  url={https://proceedings.mlr.press/v235/wang24z.html}
}

@misc{qwen3,
      title={Qwen3 Technical Report}, 
      author={An Yang and Anfeng Li and Baosong Yang and Beichen Zhang and Binyuan Hui and Bo Zheng and Bowen Yu and Chang Gao and Chengen Huang and Chenxu Lv and Chujie Zheng and Dayiheng Liu and Fan Zhou and Fei Huang and Feng Hu and Hao Ge and Haoran Wei and Huan Lin and Jialong Tang and Jian Yang and Jianhong Tu and Jianwei Zhang and Jianxin Yang and Jiaxi Yang and Jing Zhou and Jingren Zhou and Junyang Lin and Kai Dang and Keqin Bao and Kexin Yang and Le Yu and Lianghao Deng and Mei Li and Mingfeng Xue and Mingze Li and Pei Zhang and Peng Wang and Qin Zhu and Rui Men and Ruize Gao and Shixuan Liu and Shuang Luo and Tianhao Li and Tianyi Tang and Wenbiao Yin and Xingzhang Ren and Xinyu Wang and Xinyu Zhang and Xuancheng Ren and Yang Fan and Yang Su and Yichang Zhang and Yinger Zhang and Yu Wan and Yuqiong Liu and Zekun Wang and Zeyu Cui and Zhenru Zhang and Zhipeng Zhou and Zihan Qiu},
      year={2025},
      howpublished={arXiv:2505.09388},
      url={https://arxiv.org/abs/2505.09388}, 
}

@inproceedings{zhang2025physreason,
  title={PhysReason: A Comprehensive Benchmark towards Physics-Based Reasoning},
  author={Zhang, Xinyu and Dong, Yuxuan and Wu, Yanrui and Huang, Jiaxing and Jia, Chengyou and Fernando, Basura and Shou, Mike Zheng and Zhang, Lingling and Liu, Jun},
  booktitle={Proceedings of the 63rd Annual Meeting of the Association for Computational Linguistics (Volume 1: Long Papers)},
  year={2025},
  address={Vienna, Austria},
  publisher={Association for Computational Linguistics},
  pages={16593--16615},
  doi={10.18653/v1/2025.acl-long.811},
  url={https://aclanthology.org/2025.acl-long.811/}
}

@inproceedings{xu2025ugphysics,
  title={UGPhysics: A Comprehensive Benchmark for Undergraduate Physics Reasoning with Large Language Models},
  author={Xu, Xin and Xu, Qiyun and Xiao, Tong and Chen, Tianhao and Yan, Yuchen and Zhang, Jiaxin and Diao, Shizhe and Yang, Can and Wang, Yang},
  booktitle={International Conference on Machine Learning (ICML)},
  year={2025},
  url={https://openreview.net/forum?id=EmLiyZGvrR},
  note={ICML 2025 poster. arXiv:2502.00334}
}

@misc{openai_gpt52_2025,
  author={{{OpenAI}}},
  title={GPT-5.2},
  year={2025},
  howpublished={Model release},
  url={https://openai.com/index/introducing-gpt-5-2/},
  note={Large language model. Accessed 2026-01-29}
}

@misc{xai_grok_4_1_2025,
  author={{{xAI}}},
  title={Grok 4.1},
  year={2025},
  howpublished={Model release},
  url={https://x.ai/news/grok-4-1},
  note={Large language model. Accessed 2026-01-29}
}

@misc{anthropic_claude_haiku_4_5_2025,
  author={{{Anthropic}}},
  title={Claude Haiku 4.5},
  year={2025},
  howpublished={Model release},
  url={https://www.anthropic.com/news/claude-haiku-4-5},
  note={Large language model. Accessed 2026-01-29}
}

@misc{google_gemini_3_pro_2025,
  author={{{Google DeepMind}}},
  title={Gemini 3 Pro},
  year={2025},
  howpublished={Model release},
  url={https://deepmind.google/models/gemini/pro/},
  note={Large language model. Accessed 2026-01-29}
}

@misc{xai_grok_4_2025,
  author={{{xAI}}},
  title={Grok 4},
  year={2025},
  howpublished={Model release},
  url={https://x.ai/news/grok-4},
  note={Large language model. Accessed 2026-01-29}
}

@misc{anthropic_claude_sonnet_4_2025,
  author={{{Anthropic}}},
  title={Claude Sonnet 4},
  year={2025},
  howpublished={Model release},
  url={https://www.anthropic.com/news/claude-4},
  note={Large language model. Accessed 2026-01-29}
}

@misc{anthropic_claude_opus_4_2025,
  author={{{Anthropic}}},
  title={Claude Opus 4},
  year={2025},
  howpublished={Model release},
  url={https://www.anthropic.com/news/claude-4},
  note={Large language model. Accessed 2026-01-29}
}

@misc{deepseek_r1_2025,
  author={{{DeepSeek}}},
  title={DeepSeek-R1},
  year={2025},
  howpublished={Model release},
  url={https://api-docs.deepseek.com/news/news250120},
  note={Large language model. Accessed 2026-01-29}
}
\bibliographystyle{icml2026}

\newpage
\appendix
\onecolumn
   \begin{table}[htbp]
  \centering
  \caption{Distribution of scenarios in the training dataset by operator
  type and physical domain.}
  \label{tab:scenarios}
  \begin{tabular}{llcccc}
  \toprule
  \textbf{Category} & \textbf{Scenario Name} & \textbf{Operator} &
  \textbf{Sing.} & \textbf{\#Samples} \\
  \midrule
  \multirow{3}{*}{Electrostatics}
  & Point Charge Potential & Laplacian & \checkmark & 500 \\
  & Electric Dipole Potential & Laplacian & \checkmark & 500 \\
  & Coulomb Potential Field & Laplacian & \texttimes & 500 \\
  \midrule
  \multirow{6}{*}{Heat Transfer}
  & Gaussian Heat Source & Laplacian & \texttimes & 500 \\
  & Multi-Gaussian Heat Sources & Laplacian & \texttimes & 500 \\
  & Thermal Boundary Layer & Laplacian & \texttimes & 500 \\
  & Radial Heat Conduction & Laplacian & \texttimes & 500 \\
  & Anisotropic Heat Transfer & Aniso. Laplacian & \texttimes & 500 \\
  & Thermal Conduction Rings & Laplacian & \texttimes & 500 \\
  \midrule
  \multirow{4}{*}{Fluid Dynamics}
  & Potential Flow Superposition & Laplacian & \texttimes & 500 \\
  & Point Source/Sink Flow & Laplacian & \checkmark & 500 \\
  & Point Vortex Streamfunction & Laplacian & \checkmark & 500 \\
  & Doublet Flow Pattern & Laplacian & \texttimes & 500 \\
  \midrule
  \multirow{5}{*}{Quantum Mechanics}
  & Harmonic Oscillator Eigenmode & Schrödinger & \texttimes & 500 \\
  & Square Well Eigenstate & Schrödinger & \texttimes & 500 \\
  & Helmholtz Wave Function & Helmholtz & \texttimes & 500 \\
  & Bessel Scattering State & Helmholtz & \texttimes & 500 \\
  & Airy Function Solution & Schrödinger (1D) & \texttimes & 500 \\
  \midrule
  \multirow{11}{*}{Other PDEs}
  & Poisson with Poly. Source & Laplacian & \texttimes & 500 \\
  & Poisson Mixed BCs & Laplacian & \texttimes & 500 \\
  & Helmholtz Plane Wave & Helmholtz & \texttimes & 500 \\
  & Poisson with Two Point Sources & Laplacian & \checkmark & 500 \\
  & Yukawa Screened Potential & Mod. Helmholtz & \texttimes & 500 \\
  & Mod. Helmholtz Eigenmode & Mod. Helmholtz & \texttimes & 500 \\
  & Helmholtz Standing Wave & Helmholtz & \texttimes & 500 \\
  & Reaction-Diffusion Equilibrium & Nonlinear R-D & \texttimes & 500 \\
  & Fokker-Planck Stationary & Fokker-Planck & \texttimes & 500 \\
  & Line Mass Gravity Potential & Laplacian & \checkmark & 500 \\
  & Helmholtz Mode Superposition & Helmholtz & \texttimes & 500 \\
  \midrule
  \multicolumn{1}{l}{Screened Potentials}
  & Yukawa Potential & Mod. Helmholtz & \checkmark & 500 \\
  \midrule
  \multicolumn{3}{l}{\textbf{Total}} & & \textbf{15,000} \\
  \bottomrule
  \end{tabular}
  \end{table}

\newpage

\begin{figure*}[t]
    \centering
    \fcolorbox{black}{lightgray!20}{\parbox{0.95\textwidth}{
    \subsection*{Stage 1: Visual Feature Observation}

    \textbf{Purpose:} Extract visual patterns from scalar field and gradient visualizations.

    \textbf{Prompt:}
    \begin{quote}
    You are analyzing a scalar field $u(x,y)$ and its gradient components over the domain $[\text{x}_{\min}, \text{x}_{\max}, \text{y}_{\min}, \text{y}_{\max}]$.

    \textbf{Task}: Carefully observe the provided images and describe all visual features you can identify.

    \textbf{Images provided}:
    \begin{itemize}
        \item Scalar field visualization ($u(x,y)$)
        \item Gradient x-component ($\partial u/\partial x$)
        \item Gradient y-component ($\partial u/\partial y$)
    \end{itemize}

    \textbf{Instructions}:
    \begin{enumerate}
        \item Describe the overall pattern/shape of the scalar field
        \item Identify any special features: symmetries, extrema, zeros, singularities, boundaries
        \item Read values from the colorbars to estimate:
        \begin{itemize}
            \item Maximum and minimum values of $u(x,y)$
            \item Approximate values of $\partial u/\partial x$ and $\partial u/\partial y$
        \end{itemize}
        \item Note any special behaviors: radial patterns, linear gradients, oscillations, decay patterns
    \end{enumerate}

    \textbf{Output format}:
SUMMARY:
- Pattern: [concise description]

- Symmetry: [none/radial/translational/rotational/other]

- Max value: [estimate from colorbar]

- Min value: [estimate from colorbar]

- Gradient type: [uniform/varying/radial/other]

- Special features: [list key features]
    \end{quote}
    }}
    \caption{Stage 1: Visual Feature Observation Prompt for CoT Generation}
    \label{fig:stage1_prompt}
\end{figure*}

\newpage

\begin{figure*}[t]
    \centering
    \fcolorbox{black}{lightgray!20}{\parbox{0.95\textwidth}{
    \subsection*{Stage 2: Numerical Data Analysis}

    \textbf{Purpose:} Verify visual observations with quantitative evidence from CSV data.

    \textbf{Prompt:}
    \begin{quote}
    You are analyzing numerical data for a scalar field $u(x,y)$ over domain $[\text{x}_{\min}, \text{x}_{\max}, \text{y}_{\min}, \text{y}_{\max}]$.

    \textbf{Previous analysis (Stage 1)}: [Visual observations summary]

    \textbf{Task}: Analyze the provided CSV data and statistical features to extract quantitative evidence.

    \textbf{Data provided}:
    \begin{itemize}
        \item Field CSV data ($20\times20$ grid, 400 points)
        \item Gradient CSV data ($20\times20$ grid, 400 points)
        \item Statistical features:
        \begin{itemize}
            \item domain, grid\_shape
            \item extrema: max\_value, max\_location, min\_value, min\_location
            \item radial: radial\_symmetry\_score, decay\_rate, radial\_profile
            \item symmetry: x\_axis, y\_axis, rotational\_180
            \item zero\_crossings: has\_zero\_crossings, crossing\_count
            \item gradient: mean, std, range, magnitude, uniformity
            \item boundary: left, right, top, bottom values
        \end{itemize}
    \end{itemize}

    \textbf{Instructions}:
    \begin{enumerate}
        \item Verify the image observations with numerical data
        \item Calculate key numerical metrics (range, gradients, extrema locations)
        \item Identify numerical patterns (decay rates, oscillation frequencies, radial profiles)
        \item Look for quantitative clues about the solution form
    \end{enumerate}

    \textbf{Output format}:

\texttt{NUMERICAL\_EVIDENCE:}
    \begin{itemize}
        \item Max: [value] at approximately [location]
        \item Min: [value] at approximately [location]
        \item Gradient magnitude: [range]
        \item Decay/growth rate: [estimate if applicable]
        \item Oscillation frequency: [estimate if applicable]
        \item Key ratios: [any useful ratios between quantities]
    \end{itemize}

    \end{quote}
    }}
    \caption{Stage 2: Numerical Data Analysis Prompt for CoT Generation}
    \label{fig:stage2_prompt}
\end{figure*}

\newpage

\begin{figure*}[t]
    \centering
    \fcolorbox{black}{lightgray!20}{\parbox{0.95\textwidth}{
    \subsection*{Stage 3: Ground Truth Feature Extraction}

    \textbf{Purpose:} Identify all theoretical features from the analytical solution.

    \textbf{Prompt:}
    \begin{quote}
    You are analyzing the ground truth solution to identify its theoretical features.

    \textbf{Ground truth solution}: $u(x,y) = $ [analytical expression]

    \textbf{Physical/Mathematical Context}:
    \begin{itemize}
        \item PDE equation: [e.g., $(\Delta + k^2)\psi = 0$]
        \item Operator type: [e.g., Helmholtz, Laplacian, Schrödinger]
        \item Scenario: [e.g., Cylindrically Symmetric Scattering State]
    \end{itemize}

    \textbf{Task}: Extract all theoretical features from this analytical solution that could be observable in images or numerical data.

    \textbf{Instructions}:
    \begin{enumerate}
        \item Identify the solution family/type (polynomial, exponential, trigonometric, Bessel, etc.)
        \item Extract parameters and their meanings
        \item List observable features:
        \begin{itemize}
            \item Symmetries (radial, translational, rotational)
            \item Extrema locations and values
            \item Boundary behavior
            \item Decay/growth rates
            \item Oscillation patterns
            \item Zero crossings
        \end{itemize}
        \item Identify which features are most distinctive and easiest to observe
    \end{enumerate}

    \textbf{Output format}:
    
GTFEATURES:

- Solution family: [type]

- Parameters: [list with meanings]

- Observable features:

  * Feature 1: [description + how to observe]
  
  * Feature 2: [description + how to observe]
  
  ...
  
- Most distinctive features: [rank top 3]

- Verification signatures: [what would confirm this?]

    \end{quote}
    }}
    \caption{Stage 3: Ground Truth Feature Extraction Prompt}
    \label{fig:stage3_prompt}
\end{figure*}

\newpage

\begin{figure*}[t]
    \centering
    \fcolorbox{black}{lightgray!20}{\parbox{0.95\textwidth}{
    \subsection*{Stage 4: Feature Matching and Verification}

    \textbf{Purpose:} Match observed features against theoretical predictions.

    \textbf{Prompt:}
    \begin{quote}
    You are matching observed features with theoretical predictions to filter noise and confirm patterns.

    \textbf{Task}: Systematically compare observed features against theoretical predictions to identify:
    \begin{enumerate}
        \item Which GT features are clearly present in observations
        \item Which GT features are ambiguous or hard to confirm
        \item Any contradictions that need resolution
    \end{enumerate}

    \textbf{Instructions}:
    \begin{enumerate}
        \item For each GT feature, check if it appears in Stage 1 or Stage 2 observations
        \item Rate the match quality: STRONG  MODERATE  WEAK  ABSENT  CONTRADICTORY
        \item Identify which parameters can be estimated from which observations
        \item Flag any inconsistencies between images and numerical data
    \end{enumerate}

    \textbf{Output format}:
    
FEATUREMATCHING:

Confirmed features (STRONG match):
- [Feature]: [evidence from Stage 1 2]

Probable features (MODERATE match):
- [Feature]: [evidence from Stage 1 2]

Unclear features (WEAK/ABSENT):
- [Feature]: [why unclear]

Contradictions:
- [Any contradictions and potential resolutions]

PARAMETER OBSERVABILITY:

- Parameter 1: Observable from [source1, source2, ...]

- Parameter 2: Observable from [source1, source2, ...]

    \end{quote}
    }}
    \caption{Stage 4: Feature Matching and Verification Prompt}
    \label{fig:stage4_prompt}
\end{figure*}

\newpage

\begin{figure*}[t]
    \centering
    \fcolorbox{black}{lightgray!20}{\parbox{0.95\textwidth}{
    \subsection*{Stage 5: Multi-Source Parameter Estimation}

    \textbf{Purpose:} Estimate solution parameters from multiple independent sources with cross-validation.

    \textbf{Prompt:}
    \begin{quote}
    You are estimating solution parameters from multiple independent sources.

    \textbf{Task}: For each parameter identified in Stage 4, estimate its value from $\geq 2$ independent sources.

    \textbf{Instructions}:
    \begin{enumerate}
        \item For each parameter, identify 2-3 estimation methods:
        \begin{itemize}
            \item From colorbar readings
            \item From extrema values/locations
            \item From gradient magnitudes
            \item From decay rates
            \item From zero crossings
            \item From boundary values
        \end{itemize}
        \item Show explicit calculations for each method
        \item Compare estimates for consistency
        \item Compute weighted average if consistent, or flag conflicts
    \end{enumerate}

    \textbf{Example format for one parameter}:

Parameter: $\lambda$ (decay constant)

Method 1 (Colorbar):
- Center value u(0,0) $\approx$ 2.5 (from colorbar)
- If u = A*exp($-\lambda$*r), and at r=0: A $\approx$ 2.5
- At r=3, u $\approx$ 0.5 (from colorbar)
- 0.5 = 2.5*exp($-\lambda$*3) $\rightarrow$ $\lambda$ $\approx$ 0.54

Method 2 (Gradient): [similar detailed calculation]

Method 3 (Numerical): [similar detailed calculation]

Consistency: Methods agree within 8\%
Final estimate: $\lambda$ $\approx$ 0.54 $\pm$ 0.03

    \textbf{Output format}:
    
PARAMETER ESTIMATES:

Parameter 1: [value] and [uncertainty]
  (sources: [list], consistency: [\%])
  
  Parameter 2: [value] and [uncertainty]
  (sources: [list], consistency: [\%])

    \end{quote}
    }}
    \caption{Stage 5: Multi-Source Parameter Estimation Prompt}
    \label{fig:stage5_prompt}
\end{figure*}

\newpage

\begin{figure*}[t]
    \centering
    \fcolorbox{black}{lightgray!20}{\parbox{0.95\textwidth}{
    \subsection*{Stage 6: Chain-of-Thought Generation}

    \textbf{Purpose:} Generate natural reasoning from observations to solution.

    \textbf{Prompt:}
    \begin{quote}
    You are generating a Chain-of-Thought (CoT) reasoning process for predicting a PDE solution.

    \textbf{Ground Truth Solution}: [for reference, NOT to be mentioned in CoT]

    \textbf{Previous Stages Summary}: [Stage 1-5 summaries provided]

    \textbf{Task}: Generate a natural, logical Chain-of-Thought that:
    \begin{enumerate}
        \item Starts from observations (what you see in images/data)
        \item Identifies patterns and makes hypotheses about solution type
        \item Estimates parameters through explicit calculations
        \item Arrives at the final solution
        \item Verifies the solution makes sense
    \end{enumerate}

    \textbf{Critical requirements}:
    \begin{itemize}
        \item Use natural language (like human reasoning, not JSON)
        \item Show explicit arithmetic calculations
        \item Use multi-source parameter verification
        \item DO NOT say ``the ground truth is...'' or ``comparing with GT...''
        \item Make it seem like independent reasoning from observations
        \item The CoT should lead naturally to the GT solution
    \end{itemize}

    \textbf{Style reference}:

    {\small\ttfamily
<thinking>\\
Looking at the scalar field image, I observe a strong
radial symmetry centered at the origin. The colorbar
shows the field decays from approximately 2.5 at the
center to near 0 at the boundaries.\\[4pt]
Let me estimate parameters:\\
1. From colorbar: center value A $\approx$ 2.5\\
2. From decay: at r$\approx$3, u$\approx$0.5, so
   0.5 = 2.5*exp(-3$\lambda$) $\rightarrow$ $\lambda$ $\approx$ 0.536\\
3. Verification from gradient: ...\\
{[continue reasoning with calculations]}\\
</thinking>\\[4pt]
<solution>[final SymPy expression]</solution>
    }

    The CoT should be 300--800 words, showing detailed step-by-step reasoning.
    \end{quote}
    }}
    \caption{Stage 6: Chain-of-Thought Generation Prompt}
    \label{fig:stage6_prompt}
\end{figure*}

\newpage

\begin{figure*}[t]
    \centering
    \fcolorbox{black}{lightgray!20}{\parbox{0.95\textwidth}{
    \subsection*{Test Prompt: Direct Symbolic Regression from Field Data}

    \textbf{Purpose:} Evaluate model's ability to derive symbolic expressions from PDE solution data.

    \textbf{Prompt:}
    \begin{quote}
    You are an expert in analyzing scientific field data and deriving symbolic mathematical expressions.

    \section*{Input Data}

    \subsection*{1. Scalar Field Visualization}
    The first image shows the scalar field $u(x,y)$ as a heatmap.

    \subsection*{2. Gradient Components Visualization}
    The second image shows the gradient components $\partial u/\partial x$ and $\partial u/\partial y$.

    \subsection*{3. Field Data (CSV)}

    {\small\ttfamily
Data shape: (400, 3)\\
Columns: x, y, u\\
Value ranges:\\
\hspace*{1em}x: [xmin, xmax]\\
\hspace*{1em}y: [ymin, ymax]\\
\hspace*{1em}u: [umin, umax]\\[4pt]
First 10 rows:\\
\hspace*{1em}x \hspace*{2em} y \hspace*{2em} u\\
{[data rows...]}
    }

    \subsection*{4. Gradient Data (CSV)}

    {\small\ttfamily
Data shape: (400, 4)\\
Columns: x, y, du\_dx, du\_dy\\
Value ranges:\\
\hspace*{1em}x: [xmin, xmax]\\
\hspace*{1em}y: [ymin, ymax]\\
\hspace*{1em}du\_dx: [grad\_x\_min, grad\_x\_max]\\
\hspace*{1em}du\_dy: [grad\_y\_min, grad\_y\_max]\\[4pt]
First 10 rows:\\
\hspace*{1em}x \hspace*{2em} y \hspace*{1.5em} du\_dx \hspace*{1em} du\_dy\\
{[data rows...]}
    }

    \section*{Task}

    Based on the visualizations and numerical data above, derive the symbolic expression for the scalar field $u(x,y)$.

    \section*{Required Output Format}

    Your response MUST follow this exact structure:

    {\small\ttfamily
<thinking>\\
Provide your detailed reasoning process here:\\
1. Analyze the patterns, symmetries, and mathematical properties observed\\
2. Identify key features (e.g., radial symmetry, polynomial behavior, special functions)\\
3. Propose candidate symbolic expressions\\
4. Verify hypotheses against the observed data\\
5. Refine the solution based on verification results\\
</thinking>\\[4pt]
<solution>\\
Provide the final symbolic expression in SymPy format.\\
Examples:\\
- x**2 + y**2\\
- sin(x)*cos(y)\\
- exp(-x**2 - y**2)\\
- besselj(0, sqrt(x**2 + y**2))\\
</solution>
    }

    \end{quote}
    }}
    \caption{Test/Inference Prompt for Symbolic Regression Evaluation}
    \label{fig:test_prompt}
\end{figure*}

\end{document}